\documentclass[twoside,leqno,twocolumn]{article}

\usepackage[letterpaper]{geometry}

\usepackage{ltexpprt}
\usepackage{hyperref}

\usepackage{cite}
\usepackage{amsmath,amssymb,amsfonts}
\usepackage{algorithmic}
\usepackage{graphicx}
\usepackage{textcomp}
\usepackage{xcolor}
\usepackage{url}  
\usepackage{color}
\usepackage{amsmath}
\usepackage{algorithm}
\usepackage{algorithmic}
\usepackage{calc}

\usepackage{bm}
\newtheorem{problem}{Problem}

\usepackage{balance} 
\newcommand{\hide}[1]{}
\newcommand{\he}[1]{{\textsf{\textcolor{red}{[From He: #1]}}}}
\newcommand{\lc}[1]{{\textsf{\textcolor{blue}{[From Lecheng: #1]}}}}

\newcommand{\mkclean}{
  \renewcommand{\he}[1]{}
  \renewcommand{\lc}[1]{}
}
\mkclean

\newcommand{\eg}{e.g.}
\newcommand{\ie}{i.e.}

\DeclareMathOperator{\softmax}{softmax}

\DeclareMathOperator{\E}{\mathbb{E}}

\newcommand{\method}{{\textsc{Fair-MVC}}}

\def\BibTeX{{\rm B\kern-.05em{\sc i\kern-.025em b}\kern-.08em
    T\kern-.1667em\lower.7ex\hbox{E}\kern-.125emX}}

\begin{document}

\newcommand\relatedversion{}
\renewcommand\relatedversion{\thanks{The full version of the paper can be accessed at \protect\url{https://arxiv.org/abs/1902.09310}}} 

\title{Fairness-aware Multi-view Clustering}

\author{Lecheng Zheng\thanks{University of Illinois Urbana-Champaign, \{lecheng4, jingrui\}@illinois.edu}
\and Yada Zhu\thanks{MIT-IBM Watson AI Lab, IBM Researh, yzhu@us.ibm.com}
\and Jingrui He$^*$}

\date{}

\maketitle


\fancyfoot[R]{\scriptsize{Copyright \textcopyright\ 2023 by SIAM\\
Unauthorized reproduction of this article is prohibited}}

\begin{abstract}
    In the era of big data, we are often facing the challenge of data heterogeneity and the lack of label information simultaneously. In the financial domain (e.g., fraud detection), the heterogeneous data may include not only numerical data (e.g., total debt and yearly income), but also text and images (e.g., financial statement and invoice images). At the same time, the label information (e.g., fraud transactions) may be missing for building predictive models. To address these challenges, many state-of-the-art multi-view clustering methods have been proposed and achieved outstanding performance. However, these methods typically do not take into consideration the fairness aspect and are likely to generate biased results using sensitive information such as race and gender. Therefore, in this paper, we propose a fairness-aware multi-view clustering method named~\method. It incorporates the group fairness constraint into the soft membership assignment for each cluster to ensure that the fraction of different groups in each cluster is approximately identical to the entire data set. Meanwhile, we adopt the idea of both contrastive learning and non-contrastive learning and propose novel regularizers to handle heterogeneous data in complex scenarios with missing data or noisy features. Experimental results on real-world data sets demonstrate the effectiveness and efficiency of the proposed framework. We also derive insights regarding the relative performance of the proposed regularizers in various scenarios.
\end{abstract}

\begin{figure}[t]
\centering
\includegraphics[width=0.70\linewidth]{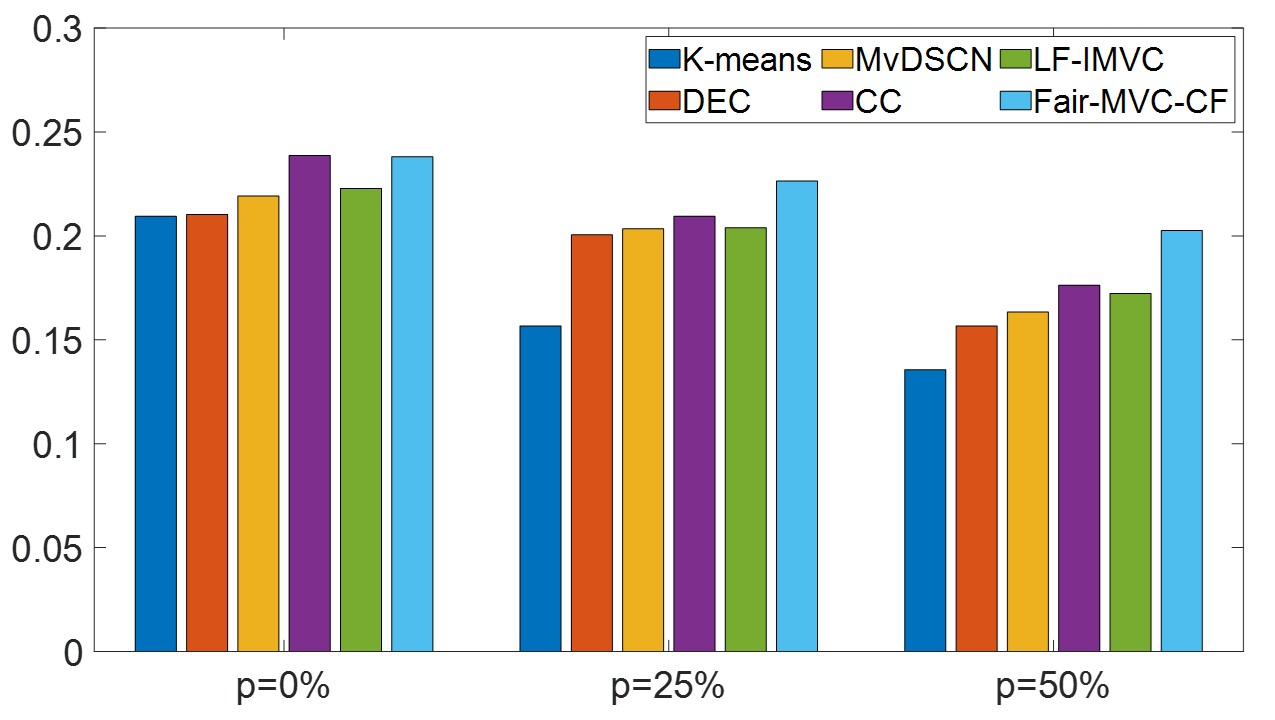} 
\vspace{-0.3cm}
\caption{Performance of SOTA methods on the Credit Card data set in missing feature scenario, where $p$ is the percentage of missing features and the y-axis is NMI score.\\}
\label{fig_fair_clustering_motivation}
\vspace{-0.3cm}
\end{figure}

\section{Introduction}
In the era of big data, the volume of data grows at an unprecedented rate. Compared with homogeneous data in the past, nowadays, the data collected from many real-world applications usually exhibit the nature of heterogeneity (\eg, view heterogeneity). For instance, on social media, one or two decades ago, users shared their daily lives with others mainly via text data; but with the development of electronic devices, users tend to share their experiences by a mixture of multiple types of data, such as a recorded video or several photos along with the text description.
Another example is in the financial domain. Take fraud detection as an example: the heterogeneous data may include not only numerical data (e.g., total debt and yearly income) but also text and images (e.g., financial statements and invoice images).
On the other hand, with the advent of big data across multiple high-impact domains, the label information is largely lacking. This phenomenon may be caused by the expensive labeling cost or the mismatch between the speed of generating data and labeling data~\cite{morden2014choosing}. Regardless of the reasons behind this phenomenon, exploring and analyzing these newly-created data is urgent in many domains~\cite{das2018real, lee2017stock}. 
To address this problem, many state-of-the-art (SOTA) multi-view clustering algorithms have been proposed, including the earliest work (\eg, Co-EM algorithms~\cite{DBLP:conf/cikm/NigamG00, DBLP:conf/icml/BrefeldS04}, Canonical Correlation Analysis-based clustering methods~\cite{DBLP:conf/icml/ChaudhuriKLS09}) and current deep learning based methods~\cite{DBLP:journals/pr/HuangKX20,  DBLP:journals/tkde/XieLQLZMWT21, DBLP:conf/www/ZhouZZLH20}.
In addition, the collected data sometimes consist of missing entries or noisy data. However, many existing SOTA algorithms~\cite{DBLP:conf/icml/ChaudhuriKLS09, DBLP:journals/pr/HuangKX20, DBLP:conf/www/ZhengCYCH21, DBLP:conf/sdm/ZhengCH19, DBLP:journals/tkde/XieLQLZMWT21,DBLP:conf/cikm/FuXLTH20} fail to effectively handle such complex scenarios. For instance, Figure~\ref{fig_fair_clustering_motivation} shows the performance of SOTA methods in terms of normalized mutual information score on the Credit Card data set~\cite{yeh2009comparisons} in the missing feature scenario. In particular, the x-axis is the percentage of the missing features and the y-axis is the normalized mutual information score (NMI). By observation, when the percentage of missing features increases, the performance of these state-of-the-art methods (\eg, DEC, LF-IMVC, CC) starts to decrease dramatically, suggesting that they couldn't effectively handle the missing feature scenario.

On the other hand, the collected data may contain sensitive information (\eg, race, gender) in many domains. The straightforward application of existing machine learning algorithms may render severely biased results~\cite{DBLP:conf/innovations/DworkHPRZ12}. 
For instance, when analyzing whether a bank should increase the interest rate for a credit card holder,
the algorithms should make predictions independent of sensitive information, such as race and gender.
In other words, these algorithms are expected to achieve good performance while satisfying the fairness constraint. Despite the outstanding performance of these aforementioned methods for addressing their respective problems~\cite{DBLP:conf/aaai/ZhaoDF17, DBLP:conf/aaai/LiJZ14, DBLP:journals/pr/HuangKX20, DBLP:conf/icml/XieGF16,  DBLP:journals/tkde/XieLQLZMWT21}, most (if not all) of these multi-view clustering methods only aim to improve the performance, and thus fail to consider the fairness constraint. Besides, though the existing fair single view clustering methods~\cite{DBLP:conf/icml/ChenFLM19, DBLP:conf/cvpr/LiZL20, DBLP:conf/icml/ZemelWSPD13} achieve the excellent performance, we couldn't directly apply them to handle multi-view data sets as a study~\cite{DBLP:journals/corr/abs-1304-5634} shows that simply concatenating multiple views into one feature vector may lead to sub-optimal solution. 

To fill in this gap, in this paper, we propose a fairness-aware multi-view clustering method named \method. It seamlessly integrates the fairness constraint into the clustering process of multi-view data. More specifically, \method ~maximizes the mutual agreement of the soft membership assignment from each view to generate the clusters. In the meanwhile, it incorporates the group fairness constraint into the soft membership assignment for each cluster to ensure that the fraction of different groups in each cluster is approximately identical to the fraction in the whole data set. In addition, to handle heterogeneous data in complex scenarios with missing data or noisy features, we adopt the idea of contrastive learning and non-contrastive learning and propose novel regularizers.

Our main contributions are summarized below.
\begin{itemize}
    \item We formalize a new problem setting: fairness-aware multi-view clustering;
    \item We propose novel contrastive and non-contrastive regularizations to handle complex scenarios with missing data or noisy features;
    \item We provide insights regarding the relative performance of contrastive and non-contrastive regularizers in various scenarios;
    \item Experimental results on both synthetic and real-world data sets demonstrate the effectiveness and efficiency of the proposed framework.
\end{itemize}
The rest of this paper is organized as follows. After a brief review of the related work in Section 2, we introduce the problem definition and our proposed framework to address this problem in Section 3. The systematic evaluation of the proposed framework on both synthetic and real-world data sets is presented in Section 4 before we conclude the paper in Section 5.

\section{Related Work}
In this section, we briefly review the related works.

\noindent\textbf{Multi-view Clustering:}
Multi-view clustering has been studied for decades. Starting from the earliest work, such as Co-EM algorithms~\cite{DBLP:conf/cikm/NigamG00, DBLP:conf/icml/BrefeldS04}, Canonical Correlation Analysis-based clustering methods~\cite{DBLP:conf/icml/ChaudhuriKLS09}, to current works~\cite{DBLP:conf/aaai/LiJZ14, DBLP:journals/corr/abs-1908-01978, liu2018late}, more and more researchers pay attention to deep multi-view clustering~\cite{DBLP:journals/pr/HuangKX20, DBLP:journals/tkde/XieLQLZMWT21} due to the great performance to handle various types of data. \cite{DBLP:conf/ijcai/LiWTGY19} proposed a novel multi-view clustering method in the adversarial setting by learning the latent representation with an auto-encoder and capturing the data distribution with adversarial training. 
However, all of these neglect the importance of fairness and to bridge the gap, we propose the fairness-aware multi-view clustering method, which incorporates group fairness into our proposed multi-view clustering algorithm.

\noindent\textbf{Fairness Machine Learning:}
Recent year has witnessed the surge of the fairness machine learning algorithms~\cite{DBLP:conf/icml/ChenFLM19, DBLP:conf/icml/AgarwalBD0W18, DBLP:conf/cvpr/LiZL20, DBLP:conf/icml/ZemelWSPD13, DBLP:conf/aistats/ZafarVGG17, DBLP:conf/nips/HardtPNS16}.~\cite{DBLP:conf/icml/ZemelWSPD13} considered both group fairness and individual fairness by encoding the input data as well as fairness constraint into a latent space and meanwhile obfuscating the membership information.
~\cite{DBLP:conf/nips/HardtPNS16} proposed a fairness measure against sensitive attributes in the classification problem to ensure equal opportunity for both protected and unprotected groups.~\cite{DBLP:conf/kdd/FeldmanFMSV15} introduced a fairness measure for classification problems and provided theoretical results to demonstrate the effectiveness of the test for disparate impact on real-world datasets. Different from these fairness algorithms, we propose a novel fairness-aware clustering algorithm in a more sophisticated setting by considering the data heterogeneity.

\noindent\textbf{Contrastive Learning:}
Contrastive learning has exhibited outstanding performance by modeling the data without supervision. 
Recent studies~\cite{DBLP:conf/nips/KhoslaTWSTIMLK20, DBLP:conf/icml/ChenK0H20, DBLP:conf/kdd/ZhengXZH22, zheng2022deeper, li2022graph, jing2022x, jing2021hdmi, feng2022adversarial, DBLP:journals/corr/abs-2210-03801}
reveal a surge of research interest in contrastive learning.~\cite{DBLP:conf/eccv/TianKI20} extended contrastive coding to a multi-view setting by maximizing the mutual information between each pair of views.~\cite{DBLP:conf/cvpr/ChenH21} addressed the drawbacks of contrastive learning-based methods by removing the negative pairs and only maximizing the similarity of positive pairs. Nevertheless, directly combining the current contrastive learning with the multi-view clustering method may lead to sub-optimal performance in some specific scenarios.
To address this issue, we propose novel contrastive and non-contrastive regularizations, which enable our proposed method to handle the perturbed data in more sophisticated scenarios.

\section{Proposed \method\ Framework}
In this section, we present our proposed Fairness-Aware Multi-view Clustering (\method) framework. We first introduce the major notation and the problem definition; then we discuss the proposed \method\ framework along with the regularization terms. Finally, we provide the overall objective function.

\subsection{Notation and Problem Definition}
In this paper, we denote $\mathcal{D} =\{\bm{X^1, X^2, ..., X^v, R}\}$ as a data set with $\bm{V}$ views and $n$ samples, where $\bm{X^i} \in \mathbb{R}^{n \times d_i}$ is the input feature matrix for the $i^{th}$ view, $\bm{R}\in \mathbb{R}^{n \times d_r}$ is the sensitive features (\eg, race, gender, etc.), $d_r$ is the dimensionality of sensitive features, and $d_i$ is the dimensionality of the input features for the $i^{th}$ view. We aim to assign the $n$ samples into $k$ clusters with the membership matrix $\bm{Q^v} \in \mathbb{R}^{n \times k}$, each represented by a centroid $\bm{\mu}_j^v \in \mathbb{R}^d, j=1,...,k$, where $d$ is the dimensionality of the centroid.
Instead of clustering these samples directly in the input space, we propose to first transform these samples with a non-linear mapping $\bm{f^v}:\bm{X^v} \rightarrow \bm{Z^v}$,  \ie, $\bm{Z^v}=\bm{f^v}(\bm{X^v})$, where $\bm{Z^v} \in \mathbb{R}^d$ is the latent representation for the $v^{th}$ view. We denote $x_i$ as the $i^{th}$ sample and $z_i$ as the hidden representation of $x_i$. 
Throughout this paper, we use $\bm{x_i^j}$ to denote the $j^{th}$ view of the $i^{th}$ sample in $\bm{X^j}$, $\bm{z_i^j}$ to denote the representation of the sample $\bm{x^j_i}$ and $\bm{r_i}$ to denote the sensitive feature of the the $i^{th}$ sample.
For ease of explanation, we only consider two views in the next few subsections, although our proposed method could be naturally extended to multiple views. With all the aforementioned notations, we are ready to formalize the fairness-aware multi-view clustering problem as follows.
\begin{problem}{\textbf{Fairness-aware Multi-view Clustering}}
    \begin{description}
      \item[Input:] a set of unlabeled data $\mathcal{D}$ along with the sensitive features $\bm{R}$ and the number of the clusters $k$.
      \item[Output:]: the membership matrix $\bm{Q}$ for each sample in $\mathcal{D}$ with the fairness constraint.
    \end{description}
\end{problem}

\subsection{Fairness-Aware Multi-view Clustering}
Following the strategy in~\cite{DBLP:conf/icml/XieGF16}, we measure the similarity between the hidden representation $\bm{z_i^v}$ and centroid $\mu_j$ as follows.
\begin{equation}
    \label{eq_1}
    \bm{q}_{ij}^v = \frac{e^{sim(\bm{z_i^v}, \bm{\mu_j^v})}}{\sum_{j'}e^{sim(\bm{z_i^v}, \bm{\mu_{j'}^v)}}}
\end{equation}
where $sim(\bm{z_i^v}, \bm{\mu_j^v})=-|\bm{z_i^v}-\bm{\mu_j^v}|^2$. Here, we denote $\bm{q}_{ij}^v$ as the element in the the $i^{th}$ row and the $j^{th}$ column of $\bm{Q^v}$. After getting the probability of the soft assignment, we could update the centroid via the formulation below:
\begin{equation}
    \label{centroid}
    \bm{\mu}_j^v = \frac{\sum_{i=1}^n \bm{q}_{ij}^v\bm{z_i^v}}{\sum_{i=1}^n \bm{q}_{ij}^v}
\end{equation}
In many real-world applications, we want the clustering results to be fair, and to not discriminate against any protected group.
For instance, when a bank makes a decision to increase the interest rate for a credit card holder, some sensitive information, (\eg, race and gender) should not be included in the algorithm but fairness measurement should be taken into consideration to ensure the fair results for its customers.
Based on the above equations, to minimize the potential bias, we follow the idea proposed in~\cite{DBLP:conf/icml/KleindessnerSAM19} that each group is approximately represented with the same fraction as in the whole data set. Given the sensitive features $\bm{R}$, the group fairness constraint could be formalized as follows.
\begin{align}
    \bm{s_j} &= \frac{\sum_{i=1}^n \sum_{v=1}^V \bm{q}_{ij}^v\bm{r_i}}{\sum_{i=1}^n \sum_{v=1}^V \bm{q}_{ij}^v},  \bm{s_D} = \frac{1}{n}\sum_{i=1}^n \bm{r_i} \nonumber\\
    L_F &= \sum_{j=1}^k \|\bm{s_j} - \bm{s_D}\|^2_2
\end{align}
where $\bm{r_i}$ is the sensitive feature of the $i^{th}$ sample in $\bm{R}$, $\bm{s_j}$ represents the weighted mean of each sensitive feature in the $j^{th}$ cluster and $\bm{s_D}$ measures the average value of each sensitive feature in the whole data set. Intuitively, minimizing $L_F$ imposes the constraint that the fraction of sensitive features in each cluster should be close to the fraction of sensitive features in the whole data set. 
Besides simply adding the fairness regularization term (\ie, $L_F$) as a regularizer, we incorporate the fairness constraint in the soft assignment to further mitigate the potential bias as follows.
\begin{equation}
    \begin{split}
    \label{eq_3}
    \bm{q}_{ij}^v &= \frac{e^{sim(\bm{z_i^v}, \bm{\mu_j^v}) + \alpha G(\bm{s}_j,  \bm{s}_D,\bm{r}_i)}}{\sum_{j'}e^{sim(\bm{z_i^v}, \bm{\mu_j^v}) + \alpha G(\bm{s}_{j'}, \bm{s}_D,\bm{r}_i)}} \\
    G(\bm{s}_j, \bm{s}_D,\bm{r}_i) &= ||\bm{s}_j - \kappa - \bm{s}_D||^2_2 - ||\bm{s}_j - \bm{s}_D||^2_2 \\
    \end{split}
\end{equation}
where $\kappa = \frac{\sum_{v=1}^V \bm{q}_{ij}^v\bm{r_i}}{\sum_{i=1}^n \sum_{v=1}^V \bm{q}_{ij}^v - \sum_{v=1}^V\bm{q}_{ij}^v}$ is the re-weighted sensitive feature of the $i$-th sample and $\alpha$ is a constant parameter balancing two terms. 
The intuition of the fairness constraint $G(\bm{s}_j, \bm{s}_D,\bm{r}_i)$ is straightforward. If $G(\bm{s}_j, \bm{s}_D,\bm{r}_i)>0$, it means that removing the $i$-th sample from $j$-th cluster (\ie, $||\bm{s}_j - \kappa - \bm{s}_D||^2_2$) increases the difference between $\bm{s}_j$ and $\bm{s}_D$, and it will cause the clustering results to be unfair. Thus, we should keep the $i$-th sample in $j$-th cluster. Otherwise, we should remove the $i$-th sample from $j$-th cluster to decrease the difference. After mitigating the bias in the soft assignment, we propose to iteratively refine the clusters by minimizing the distance between $\bm{z_i}^v$ and $\bm{\mu}_i^v$ as follows.
\begin{align}
    L_d & = \sum_{i, j, v} \bm{c}_{ij}^v \|\bm{z_i^v}-\bm{\mu_j^v}\|^2_2
\end{align}
where $\bm{c}_{ij}^v\in\{0, 1\}$ denotes whether the $i^{th}$ sample belongs to the $j^{th}$ cluster based on the $v{^{th}}$ view.
$L_d$ aims to ensure that the samples belonging to the same cluster will get closer. In addition, based on the assumption~\cite{DBLP:journals/corr/abs-1304-5634} in multi-view learning that the information contained in each view is consistent, we aim to match the soft assignment made by the first view to the soft assignment made by the second view by minimizing the KL divergence between two distributions:
\begin{align}
    L_{KL} & = KL(\bm{Q^1}||\bm{Q^2}) + KL(\bm{Q^2}||\bm{Q^1}) \nonumber\\
    &= \sum_i\sum_j(\bm{q}_{ij}^1 \log \frac{\bm{q}_{ij}^1}{\bm{q}_{ij}^2} + \bm{q}_{ij}^2 \log \frac{\bm{q}_{ij}^2}{\bm{q}_{ij}^1})
\end{align}
where $\bm{Q^1}$ and $\bm{Q^2}$ are two soft assignment matrices. 

\subsection{Regularization}
The main idea of unsupervised contrastive loss is to utilize the rich unlabeled data to enhance the quality of the hidden representation. Rather than directly imposing the contrastive constraint on the latent space $\bm{Z}$, we first transform $\bm{Z}$ into another space $\bm{H}$ with the second encoder $g^v$ (\eg, $\bm{h_i^v}=g^v(\bm{z_i^v})$) by following the idea proposed in ~\cite{DBLP:conf/icml/ChenK0H20} to avoid distorting the hidden representation $\bm{Z}$ and then we regularize the hidden space $\bm{H}$ as follows.
\begin{equation}
    \label{l1}
    \begin{split}
         L_1 & = -\E_{x_i\in \mathcal{D}}[  \log \frac{f(\bm{h_i^1}, \bm{h_i^2})}{f(\bm{h_i^1}, \bm{h_i^2}) + \sum_{\bm{x_{j}} \in \mathcal{N}_{i}^\mathcal{D}} \sum_{v} f(\bm{h_i^v}, \bm{h_j^v})} ]
    \end{split}
\end{equation}
where $\bm{x_j^v}$ is the $v^{\textrm{th}}$ view of $\bm{x_j}$, $\bm{h_j^v}$ is the hidden representation of $\bm{x_j^v}$ after non-linear mappings, $f(\bm{h_i^1}, \bm{h_i^2})$ is a similarity measurement function, \eg, $f(a, b)=\exp(\frac{a \cdot b}{\tau})$, $\tau$ is the temperature, 
and $\mathcal{N}_{i}^\mathcal{D} =\mathcal{D} \backslash \{i\}$. 
However, $L_1$ suffers from the class collision problem~\cite{DBLP:journals/corr/abs-2110-04770}, where minimizing $L_1$ pushes two samples from the same cluster away from each other and thus leads to sub-optimal performance. 
To alleviate these potential concerns, we propose a novel weighting strategy as follows.
\begin{align}
\scalebox{0.82}{
    \label{l_ctr}
    $L_{ctr} = -\E_{x_i\in \mathcal{D}}[  \log \frac{f(\bm{h_i^1}, \bm{h_i^2})}{f(\bm{h_i^1}, \bm{h_i^2}) + \sum_{\bm{x_{j}} \in \mathcal{N}_{i}^\mathcal{D}}  \sum_{v} sim(\bm{q_i}, \bm{q_j}) f(\bm{h_i^v}, \bm{h_j^v})} ]$}
\end{align}
where $\bm{q_i}=[q_i^1; ... ;q_i^v]$ is the concatenation of the $v$ views soft membership for the $i^{th}$ samples and $sim(\bm{q_i}, \bm{q_j})=\exp(1 - \frac{\bm{q_i} \cdot \bm{q_j}}{|\bm{q_i}| |\bm{q_j}|})$. The intuition of the weighting function $sim(\bm{q_i}, \bm{q_j})$ is that if two samples have similar probabilities of being assigned to the same cluster, then this pair of samples should be considered as a positive pair, and we need to reduce the weight of this pair of samples in the denominator in Equation~\ref{l_ctr} in order to address the class collision issue. Notice that if $\bm{q_i}$ and $\bm{q_j}$ are equal in the extreme case, then the value of the weighting function is 1. The more dissimilar $\bm{q_i}$ and $\bm{q_j}$ are, the large the value of the weighting function $sim(\bm{q_i}, \bm{q_j})$ is.  
Equation~\ref{l1} assigns the equal weight to all negative samples, which inevitably pushes two samples from the same cluster away from each other, while in our proposed weighted contrastive loss $L_{ctr}$, we utilize the pseudo-label to alleviate such an issue. 

One drawback of contrastive learning-based regularization is the high computational cost as well as the high memory requirement to compute and store the similarity matrix for any pairs of two samples~\cite{DBLP:conf/cvpr/ChenH21}. To address this issue, a non-contrastive learning based method~\cite{DBLP:conf/cvpr/ChenH21} has been proposed:

\begin{align}
    \scalebox{0.92}{
    \label{l2}
         $L_2 = -\E_{x_i \in \mathcal{D}}(\frac{\bm{h_i^1}}{|\bm{h_i^1}|_2}\cdot SG(\frac{ \bm{z_i^2}}{|\bm{z_i^2}|_2}) +  \frac{\bm{h_i^2}}{|\bm{h_i^2}|_2} \cdot SG(\frac{ \bm{z_i^1}}{|\bm{z_i^1}|_2}))$}
\end{align}
where $SG$ denotes stop gradient operation, and $\bm{H^v_i}=g(\bm{Z^v_i}) \in \mathbb{R}^{n \times d}$. Notice that different from contrastive regularization $L_{ctr}$, $g(\cdot)$ is shared by two views in $L_2$. Intuitively, $L_2$ aims to maximize the similarity of the hidden representations of two views. However, in practice, if parts of the original features are missing or noisy, $L_2$ might also result in a sub-optimal solution, which is examined in the case study in Section~\ref{case_study}. Inspired by~\cite{DBLP:conf/nips/LuYBP16}, we propose a cross-attention module to borrow the information from the other view to alleviate this issue:
\begin{equation}
    \begin{split}
         \bm{C}_{1,2} &=\bm{H}^1 \bm{W}_{1,2} (\bm{H}^2)^T \\
         \bm{O}^1 &= \tanh(\bm{Z}^1 \bm{W}_1 + \bm{Z}^2\bm{W}_2\bm{C}_{1,2}) \\
         \bm{O}^2 &= \tanh(\bm{Z}^2 \bm{W}_2 + \bm{Z}^1\bm{W}_1\bm{C}_{1,2}^T) \\
         \bm{A}^v &= \softmax(\bm{O}^v) \\
         \bm{T}^v &= \bm{H}^v \odot \bm{A}^v \\ 
    \end{split}
\end{equation}
where $\bm{H^v}=g(\bm{Z^v})\in \mathbb{R}^{n \times d}$ denotes the hidden representation after the mapping function $g(\cdot)$, $\bm{W}_{1,2} \in \mathbb{R}^{n \times n}$, $\bm{W}_1 \in \mathbb{R}^{d \times d}$ and $\bm{W}_2 \in \mathbb{R}^{d \times d}$ are the weight matrices and $\bm{C}_{1,2} \in \mathbb{R}^{d \times d}$ aims to capture the relatedness of features across two views. By leveraging the consensus information to measure the importance of each feature, $\bm{O}^1 \in \mathbb{R}^{n \times d}$ and $\bm{O}^2 \in \mathbb{R}^{n \times d}$ encode the information from both views in order to alleviate the issue of missing or noisy features. $\bm{A}^v \in \mathbb{R}^{n \times d}$ is the attention matrix for the $v^{th}$ view, $\bm{T}^v \in \mathbb{R}^{n\times d}$ is the output of the cross attention module for the $v^{th}$ view and $\odot$ denotes the element-wise multiplication operation. 
The main difference between $\bm{H}^v$ and $\bm{T}^v$ is that $\bm{A}^v$ first encodes the information from both views and then adjusts the importance of the features in $\bm{H}^v$ based on the consensus information from both views to mitigate the issue of the missing or noisy features. 
Similar to $L_2$, the non-contrastive learning loss could be updated as follows.
\begin{align}
\scalebox{0.86}{
    \label{l_nctr}
    $L_{nctr} = -\E_{X_i \in \mathcal{D}}(\frac{\bm{t_i^1}}{|\bm{t_i^1}|_2}\cdot SG(\frac{ \bm{z_i^2}}{|\bm{z_i^2}|_2}) +  \frac{\bm{t_i^2}}{|\bm{t_i^2}|_2} \cdot SG(\frac{ \bm{z_i^1}}{|\bm{z_i^1}|_2}))$}
\end{align}

\subsection{Objective Function and Proposed Algorithm}
Now, we are ready to introduce the overall objective function:
\begin{equation}
    \label{overall}
    \begin{split}
        \min J &=  L_{KL} + \gamma L_d  + \alpha L_F + \beta L_{reg}
    \end{split}
\end{equation}
where $L_{KL}$ is KL-divergence maximizing the mutual agreement of soft assignment of two views, $L_d$ ensures that the samples belonging to the same cluster will get closer,  $L_F$ is the group fairness constraint, $L_{reg}$ is either $L_{ctr}$ or $L_{nctr}$ regularizing the latent representations, and $\alpha$, $\beta$, and $\gamma$ are positive hyper-parameters balancing these terms. Notice that $\alpha$ is the same parameter as in Equation~\ref{eq_3}. 
The proposed method could be solved in Expectation-Maximization (EM) steps. Our algorithm is presented in Algorithm
~\ref{alg1}.
Specifically, we take the results of K-means as the initial centroids of $k$ clusters in the first step. Next, we first compute the soft assignment based on Equation~\ref{eq_3}, and maximize the mutual agreement of soft membership of multiple views based on Equation~\ref{overall} in Step 2 and Step 3; then we update the centroid of each cluster based on Equation~\ref{centroid} in step 4. These steps are repeated $T$ times, where $T$ is the number of iterations. Finally, we compute the soft assignment based on Equation~\ref{eq_1} by excluding the sensitive features at the test phase.


\begin{algorithm}
    \begin{algorithmic}
        \caption{\method\ Algorithm}
        \label{alg1}
        \REQUIRE  The total number of iterations $T$, the input data ${\bm{X}^1, \bm{X}^2, ..., \bm{X}^v}$, the sensitive features $\bm{R}$ and the number of cluster $k$.\\
        \ENSURE The membership matrix $\bm{Q}$.\\
        \STATE \textbf{Step 1: } Take the output of K-means as the initial centroids of $\bm{k}$ clusters or randomly initialize the centroids.
        \FOR{$t=1$ to $T$}
            \STATE \textbf{Step 2: } Compute the soft assignment based on Equation~\ref{eq_3}.
            \STATE \textbf{Step 3: } Minimize the overall objective function based on Equation~\ref{overall}.
            \STATE \textbf{Step 4 :} Update the centroids based on Equation~\ref{centroid}.
        \ENDFOR
         \STATE \textbf{Step 5: } Compute the membership by averaging the soft assignment of different views based on Eq.~\ref{eq_1}.
    \end{algorithmic}
\end{algorithm}
\section{Experiments}
In this section, we demonstrate the performance of our proposed framework in terms of effectiveness by comparing it with state-of-the-art methods.

\subsection{Experimental Setup}

We mainly evaluate our proposed algorithm on three data sets with fairness constraints, including Credit Card data set~\cite{DBLP:conf/aistats/ZafarVGG17}, Bank Marketing data set~\cite{DBLP:conf/aistats/ZafarVGG17} and Zafar data set~\cite{DBLP:conf/aistats/ZafarVGG17}, and two data sets without fairness constraints, including Noisy MNIST~\cite{westbury1994x} and X-ray Microbeam (XRMB)~\cite{westbury1994x}.
Specifically, the sensitive feature on the Credit Card data set is gender, and the sensitive feature on the Bank Marketing data set is marital status. Zafar data set~\cite{DBLP:conf/aistats/ZafarVGG17} is a widely-used synthetic data set, where one binary value is generated as the sensitive feature. More details of these data sets could be found in Appendix 
~\ref{FairMVC_setup}. 

\noindent\textbf{Baselines:} In the experiment, $L_{ctr}$ is the regularization term (\ie, $L_{reg}$) in~\method-C and $L_{nctr}$ is the regularization term (\ie, $L_{reg}$) in \method-N. We compare the performance of our methods with the following baselines: (1). K-means: a method aiming to partition samples into several clusters where each sample is assigned to the nearest cluster; (2). DEC~\cite{DBLP:conf/icml/XieGF16}: a deep embedded clustering method by learning feature representations and cluster assignments with deep neural networks; (3). Contrastive-Clustering (CC)~\cite{DBLP:conf/aaai/Li0LPZ021}: a contrastive learning-based clustering method, which optimizes the instance-level and cluster-level contrastive loss simultaneously; (4). MvDSCN~\cite{DBLP:journals/corr/abs-1908-01978}: a multi-view deep subspace clustering network aiming to learn a multi-view self-representation matrix; (5). LF-IMVC~\cite{liu2018late}: an incomplete multi-view clustering method (as this method is designed for missing feature scenarios, we only report its performance in table \ref{table:case_study_missing_feature}). To investigate the contributions of different parts of \method-N and \method-C, we conduct an ablation study by introducing four variations of \method, including \method-NF that removes fairness constraint from \method-N, \method-CF that removes fairness constraint from \method-C, \method-1 where $L_{reg}$ is replaced by $L_1$, and \method-2 where $L_{reg}$ is replaced by $L_2$.

\noindent\textbf{Evaluation:} We present the results regarding the following metrics: (1) NMI~\cite{DBLP:books/daglib/0016881}: normalized mutual information, which measures the mutual dependency of two variables. (2) Balance: a group fairness measurement, which is formulated as follows:
\begin{align}
    Balance = \min_i\frac{\min_a|C_i\cup r_j|}{|C_i|} 
\end{align}
where $C_{i}\in\{0, 1\}$ denotes the $i$-{th} cluster and $r_j$ denotes the $j$-th protected subgroup. Typically, the upper bound of balanced is determined by the distribution of the sensitive feature, and a higher value of balance indicates a fairer result. 
\textbf{The efficiency analysis and parameter analysis could be found in 
Appendix~\ref{FairMVC_efficiency} and ~\ref{FairMVC_parameter_analysis}, respectively.}

\begin{table*}[t]
\centering
\caption{Results on three data sets with sensitive features. (Higher balance score indicates better fairness.)}
\resizebox{\columnwidth * 2}{!}{%
\begin{tabular}{|*{7}{c|}}
\hline - & \multicolumn{2}{c|}{Credit Card} & \multicolumn{2}{c|}{Zafar} & \multicolumn{2}{c|}{Bank Marketing} \\
\hline \textbf{Model}       & \textbf{NMI}     & \textbf{Balance}          & \textbf{NMI}     & \textbf{Balance}  & \textbf{NMI}     & \textbf{Balance}  \\
\hline K-means                  & 0.2094 $\pm$ 0.0114 & 0.3553 $\pm$ 0.0037 & 0.7032 $\pm$	0.0078 &	0.1706 $\pm$	0.0076  &  0.2867 $\pm$ 	0.0144 & 	0.3765 $\pm$ 	0.0066 \\   
\hline DEC                      & 0.2103 $\pm$ 0.0209 & 0.3596 $\pm$ 0.0060 & 0.7255 $\pm$	0.0192 &	0.1685 $\pm$	0.0073  &  0.3093 $\pm$ 	0.0115 & 	0.3760 $\pm$ 	0.0096   \\   
\hline MvDSCN                   & 0.2192 $\pm$ 0.0153 & 0.3582 $\pm$ 0.0041 & 0.7691 $\pm$	0.0042 &	0.1713 $\pm$	0.0065  &   0.3624 $\pm$ 	0.0055 & 	0.3759 $\pm$ 	0.0067 \\  
\hline CC                       & 0.2387 $\pm$ 0.0128 & 0.3574 $\pm$ 0.0047 & 0.7895 $\pm$	0.0068 &	0.1701 $\pm$	0.0071  &   0.3623 $\pm$ 	0.0101 & 	0.3746 $\pm$ 	0.0097 \\
\hline \method-1                & 0.2423 $\pm$ 0.0070 & 0.3743 $\pm$ 0.0028 & 0.7878 $\pm$	0.0101 &	0.2735 $\pm$	0.0077  &   0.3587 $\pm$ 	0.0050 & 	0.4116 $\pm$ 	0.0073  \\  
\hline \method-2                & 0.2434 $\pm$ 0.0039 & 0.3710 $\pm$ 0.0036 & 0.7996 $\pm$	0.0110 &	0.2767 $\pm$	0.0055  &  0.3632 $\pm$ 	0.0069 & 	0.4106 $\pm$ 	0.0092  \\  
\hline \method-CF               & 0.2386 $\pm$ 0.0100 & 0.3599 $\pm$ 0.0028 & 0.7994 $\pm$	0.0107 &	0.1770 $\pm$	0.0044  &  0.3861 $\pm$ 	0.0103 & 	0.3735 $\pm$ 	0.0070 \\   
\hline \method-NF               & \textbf{0.2484 $\pm$ 0.0095} & 0.3618 $\pm$ 0.0034 & \textbf{0.8161 $\pm$	0.0157} &	0.1768 $\pm$	0.0053 &   \textbf{0.3899 $\pm$ 	0.0091} & 	0.3776 $\pm$ 	0.0091 \\   
\hline \method-C                & 0.2459 $\pm$ 0.0078 &\textbf{ 0.3783 $\pm$ 0.0032} &  0.7974 $\pm$	0.0051 &	\textbf{0.2896 $\pm$	0.0059}  &  0.3816 $\pm$ 	0.0116 & 	0.4208 $\pm$ 	0.0059\\   
\hline \method-N               & 0.2471 $\pm$ 0.0041 & 0.3743 $\pm$ 0.0029 & 0.8119 $\pm$	0.0150 &	0.2827 $\pm$	0.0074  &  0.3839 $\pm$ 	0.0109 & 	 \textbf{0.4240 $\pm$ 	0.0075} \\  
\hline
\end{tabular}
}
\label{table:fairness}
\end{table*}

\subsection{Experimental results}
In this subsection, we demonstrate the effectiveness of the proposed method.  
Table~\ref{table:fairness} shows the performance of state-of-the-art methods and our proposed methods. By observations, we find that (1) most baselines fail to provide fair results, though many of them achieve outstanding performance; for instance, though CC achieves competitive performance on the Credit Card data set, Zafar data set, and Bank Marketing data set, its balance score is much lower than that of \method-C and \method-N; (2) \method-NF outperforms all baselines on the Credit Card data set, Zafar data set and Bank Marketing data set in Table~\ref{table:fairness} without considering the fairness; (3) comparing with state-of-the-art methods, \method-C and \method-N achieve much better balance score by taking fairness into considerations. The experimental results on non-fairness data sets (\ie, Noisy MNIST and XRMB) could be found in Appendix 
~\ref{appendix_no_fairness}.

\noindent\textbf{Ablation study} 
To demonstrate the effectiveness of each component in our proposed framework, we conduct an ablation study. In Table~\ref{table:fairness}, comparing the performance of \method-C with \method-CF, the balance score of \method-C increases by more than 63\% on the Zafar data set while the NMI of \method-C only decreases by less than 0.25\% on Zafar data set, which demonstrates the effectiveness of \method-C. By comparing the performance of \method-2 and \method-N, we demonstrate that our proposed novel non-contrastive regularizer can improve performance by leveraging information from the complementary view to some extent in the presence of the missing feature scenario. What's more, \method-C outperforms CC and \method-1 on the Credit Card data set, Zafar data set and Bank Marketing data set. As we mentioned early, the main drawback of vanilla contrastive regularizer (\ie, $L_1$ in Equation~\ref{l1}) is that minimizing the loss function pushes two samples from the same cluster away from each other, resulting in the class collision problem. \method-C and \method-CF consider the soft membership by reducing the weights for any pairs of samples possibly from the same cluster in the denominator of Equation~\ref{l_ctr}.

\begin{figure}[t]
\caption{Fairness analysis on the Credit Card data set. The y-axis is the number of males for each method. \textbf{Left}: Five bars in each group (algorithm) denote the number of males (sensitive feature) in five clusters. The more discrepancy to ground truth, the worse fairness. \textbf{Right}: each bar means the standard deviation. The more similar to the ground truth, the fairer the clustering results.}
\vspace{-0.3cm}
\begin{center}
\begin{tabular}{cc}
\hspace{-0.6cm}
\includegraphics[width=0.52\linewidth]{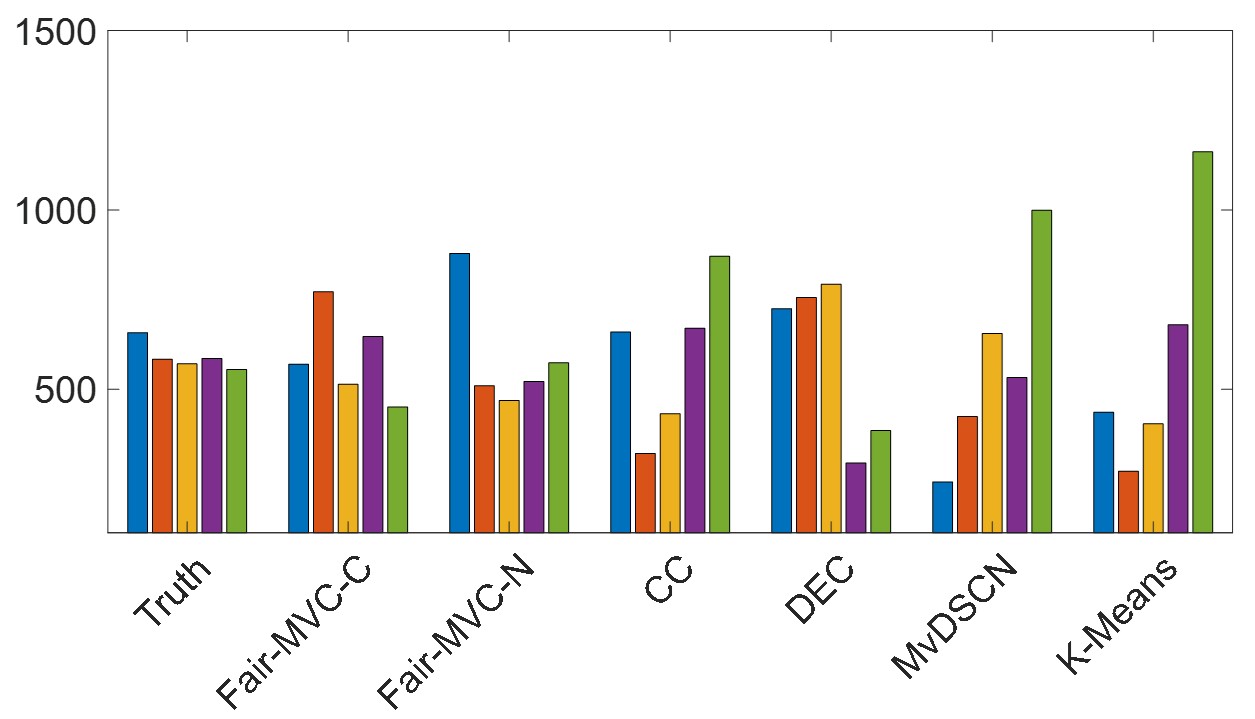} &
\includegraphics[width=0.52\linewidth]{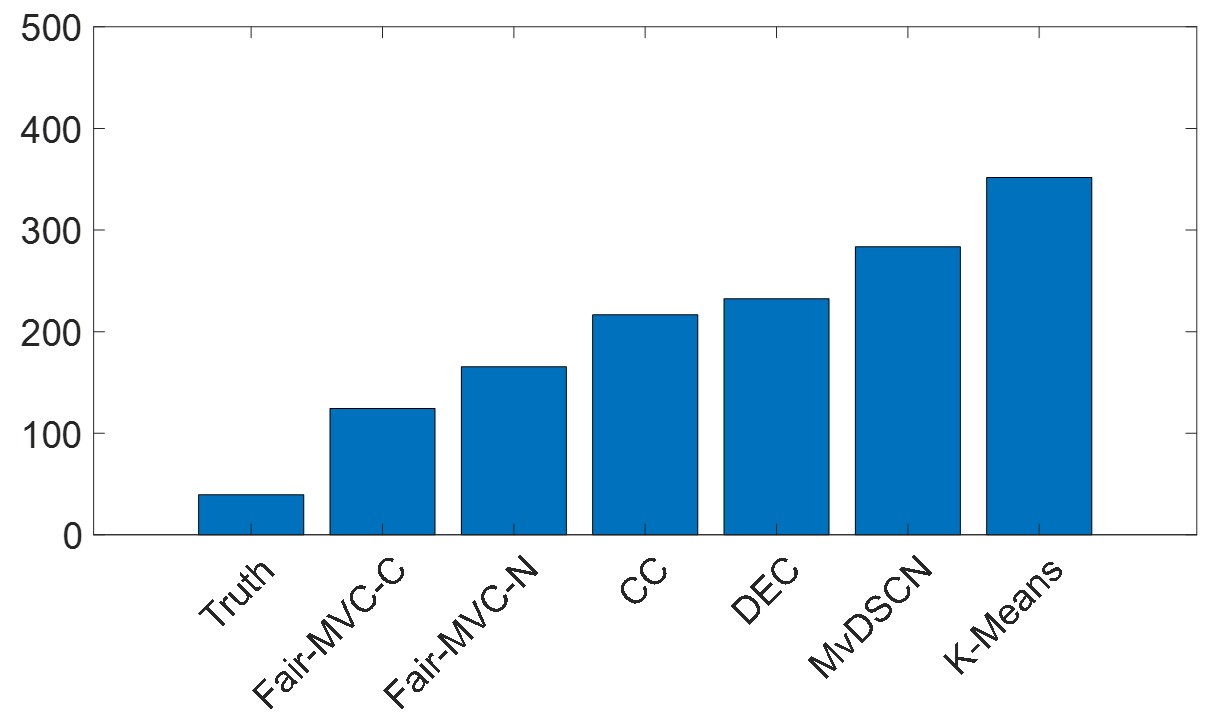} \\
\end{tabular}
\end{center}
\vspace{-0.4cm}
\label{fig_fairness_analysis}
\end{figure}

\subsection{Fairness Analysis}
\emph{Why do we care about the fairness of the clustering results?} To answer this question, let us first look at the clustering results on the Credit Card data set.  On the Credit Card data set, the attributes consist of the historical payments; the sensitive feature is gender; it consists of five clusters.
The first cluster means that the customers pay their debt duly and the rest four clusters mean that the customers fail to pay the debt in one, two, three, or more than three consecutive months. If the banks aim to determine whether to lower the interest rate of the customers based on their payment records, they want the decisions made on the clustering results to be fair and not discriminate against any protected group. Thus, reducing the potential bias is crucial for the clustering methods.
In Figure~\ref{fig_fairness_analysis}, we visualize the fairness measurement in terms of the count of males in each cluster on the Credit Card data set, as the sensitive feature in this data set is gender. In Figure~\ref{fig_fairness_analysis}, five bars (in the left figure) in each group (algorithm) denote the number of males (sensitive feature) for five clusters and the bar in the right figure means the standard deviation of the number of males for each method. Intuitively, the distribution of males for a fair clustering result should be identical to the distribution of males using the ground truth. The more dissimilar to the ground truth, the more unfair the clustering results. By observation, we find that \method-C is mostly identical to the ground truth, compared with other baselines in terms of the count distribution and the standard deviation of the count of males. These baselines fail to consider the fairness constraint, thus leading to lower balance scores.

\begin{table*}[t]
\centering
\caption{\textbf{Case Study: Missing Feature Scenario.} Results on Credit Card data set, where $p$ denotes the percentage of missing features. (Higher balance score indicates better fairness.) }
\resizebox{\columnwidth * 2}{!}{%
\begin{tabular}{|*{7}{c|}}
\hline - & \multicolumn{2}{c|}{$p=0\%$} & \multicolumn{2}{c|}{$p=25\%$} & \multicolumn{2}{c|}{ $p=50\%$} \\
\hline \textbf{Model}       & \textbf{NMI}     & \textbf{Balance}          & \textbf{NMI}     & \textbf{Balance}  & \textbf{NMI}     & \textbf{Balance}  \\
\hline K-means                  & 0.2094 $\pm$ 0.0114 & 0.3553 $\pm$ 0.0037 & 
0.1567 $\pm$	0.0148  &	0.3602 $\pm$	0.0060  & 0.1356 $\pm$ 	0.0063 &	0.3632 $\pm$	0.0038  \\   
\hline DEC                      & 0.2103 $\pm$ 0.0209 & 0.3596 $\pm$ 0.0060 & 0.2005 $\pm$	0.0079  &	0.3640 $\pm$	0.0078  &  0.1567 $\pm$	0.0121 &	0.3626 $\pm$	0.0040   \\   
\hline MvDSCN                   & 0.2192 $\pm$ 0.0153 & 0.3582 $\pm$ 0.0041 & 0.2034 $\pm$	0.0159  &	0.3594 $\pm$	0.0048  &   0.1634 $\pm$	0.0183 &	0.3663 $\pm$	0.0069 \\  
\hline CC                       & 0.2387 $\pm$ 0.0128 & 0.3574 $\pm$ 0.0047 & 0.2095 $\pm$	0.0094  &	0.3616 $\pm$	0.0047  &  0.1762 $\pm$	0.0175 &	0.3680 $\pm$	0.0049  \\
\hline LF-IMVC                  & 0.2228 $\pm$ 0.0093 &0.3625 $\pm$	0.0043 & 0.2039 $\pm$	0.0083  &	0.3581 $\pm$	0.0040  & 0.1723 $\pm$	0.0144 &	0.3621 $\pm$	0.0031 \\
\hline \method-C                & 0.2459 $\pm$ 0.0078 &\textbf{ 0.3783 $\pm$ 0.0032} &  0.2165 $\pm$	0.0054  &	\textbf{ 0.3871 $\pm$	0.0046}  &   \textbf{0.1871 $\pm$	0.0079} &	\textbf{0.3907 $\pm$	0.0076} \\   
\hline \method-N                & \textbf{0.2471 $\pm$ 0.0041} & 0.3743 $\pm$ 0.0029 &  \textbf{0.2209 $\pm$	0.0075}  &	0.3820 $\pm$	0.0121  &   0.1803 $\pm$	0.0060 &	0.3854 $\pm$	0.0114 \\  
\hline
\end{tabular}
}
\label{table:case_study_missing_feature}
\vspace{-0.3cm}
\end{table*} 

\begin{table*}[t]
\caption{\textbf{Case Study: Noisy Feature Scenario.} Results on Credit Card data set, where $p$ denotes the percentage of perturbed features (Higher balance score indicates better fairness.)}
\centering
\resizebox{\columnwidth * 2}{!}{%
\begin{tabular}{|*{7}{c|}}
\hline - & \multicolumn{2}{c|}{$p=0\%$} & \multicolumn{2}{c|}{$p=25\%$} & \multicolumn{2}{c|}{ $p=50\%$} \\
\hline \textbf{Model}       & \textbf{NMI}     & \textbf{Balance}          & \textbf{NMI}     & \textbf{Balance}  & \textbf{NMI}     & \textbf{Balance}  \\
\hline K-means                  & 0.2094 $\pm$ 0.0114 & 0.3553 $\pm$ 0.0037 & 0.1793  $\pm$ 	0.0078 & 0.3589 $\pm$  0.0080 & 0.1663 $\pm$ 0.0098 & 0.3561 $\pm$ 0.0072  \\  
\hline DEC                      & 0.2103 $\pm$ 0.0209 & 0.3596 $\pm$ 0.0060 & 0.1923  $\pm$ 	0.0087 & 0.3561 $\pm$  0.0072 & 0.1779 $\pm$ 0.0100 & 0.3617 $\pm$ 0.0030   \\   
\hline MvDSCN                   & 0.2192 $\pm$ 0.0153 & 0.3582 $\pm$ 0.0041 &  0.2011  $\pm$ 	0.0070 & 0.3612 $\pm$  0.0051 & 0.1885 $\pm$ 0.0051 & 0.3626 $\pm$ 0.0039 \\  
\hline CC                       & 0.2387 $\pm$ 0.0128 & 0.3574 $\pm$ 0.0047 & 0.2078  $\pm$ 	0.0078 & 0.3609 $\pm$  0.0087 & 0.1960 $\pm$ 0.0083 & 0.3583 $\pm$ 0.0040  \\  
\hline \method-C                & 0.2459 $\pm$ 0.0078 &\textbf{ 0.3783 $\pm$ 0.0032} &  \textbf{0.2490  $\pm$ 	0.0147} & \textbf{0.3785 $\pm$  0.0041} & \textbf{0.2397 $\pm$ 0.0052} & 0.3819 $\pm$ 0.0067 \\   
\hline \method-N                & \textbf{0.2471 $\pm$ 0.0041} & 0.3743 $\pm$ 0.0029 &  0.2394  $\pm$ 	0.0051 & 0.3741 $\pm$  0.0047 & 0.2256 $\pm$ 0.0068 & \textbf{0.3832 $\pm$ 0.0054} \\  
\hline
\end{tabular}
}
\label{table:case_study_perturbed_feature}
\vspace{-0.3cm}
\end{table*}

\subsection{Case studies: }
\label{case_study}

{\bf Missing Features.} In this subsection, we first demonstrate the effectiveness of the two regularizers in our proposed methods (\ie, \method-N and \method-C) and state-of-the-art methods
in the presence of missing features. To control the percentage of missing features $p$, we randomly mask the features with Bernoulli distribution (where $p$ is the possibility of being masked) on the Credit Card data set. Table~\ref{table:case_study_missing_feature} shows the performance of these methods and we gradually increase the ratio of missing features from 0 to 25\% and then to 50\%. Notice that the upper bound of the balance score is determined by the distribution of sensitive features, which is 0.4092 for the Credit Card data set.
Based on the results from Table~\ref{table:case_study_missing_feature}, we have the following observations: (1) when there are no missing features (\ie, $p=0\%$), \method-N outperforms \method-C on the Credit card data set; (2) when we gradually increase the percentage of missing features, \method-C gradually outperforms \method-N; (3) the performance of the most baseline methods decreases dramatically as the percentage of missing feature increases. As for the second observation, we conjecture that the contrastive regularizer maximizes the similarity between two views from the same instance, and meanwhile, it contrasts the difference between two views from two different instances. This contrasting operation leverages the information from other instances to infer the missing features, thus enhancing the quality of hidden representation. Different from the contrastive regularizer, the non-contrastive regularizer only aims to maximize the similarity between two views from the same instance, and thus it fails to infer extra information from other instances. Therefore, the performance of \method-N is a little bit worse than \method-C. 

\noindent{\bf Noisy Features.} Next, we further investigate the effectiveness of the two regularizers (\ie, non-contrastive regularizer and contrastive regularizer) in the presence of noisy features. Table~\ref{table:case_study_perturbed_feature} shows the performance of state-of-the-art methods. Here is the procedure to perturb the raw data. We first use the Bernoulli distribution to select $p$ percent of data and then inject the white noise (\eg, $\mathcal{N}(0, 1)$) into the raw data on the Credit Card data set. By observations, we find that when $p$ percent of white noise is added to raw data, the performance of most methods begins to decrease. Different from most baseline methods, the performance of \method-C decreases slightly, when 50\% percent of noise is added. We conjecture that our proposed contrastive regularizer is more robust to the noisy feature as it can contrast one instance's representation with others' representations. However, the vanilla contrastive-based method (\ie, CC) suffers from the class collision issue and two non-contrastive regularizers fail to leverage the information from other instances.

\subsection{Discussion}
Combining the experimental results from Table~\ref{table:fairness}, Table~\ref{table:case_study_missing_feature} and Table~\ref{table:case_study_perturbed_feature}, we observe that for the clean input data, the non-contrastive regularizer tends to have better performance than the contrastive regularizer for clustering, as the contrastive regularizer usually suffers from the class collision issue. Nevertheless, in the presence of missing or noisy features, we observe that the performance of the non-contrastive regularizer decreases by a large margin, while the performance of the contrastive regularizer decreases slowly by contrasting with other instances and inferring extra information from other instances. On the other hand, based on the experimental results in the efficiency analysis in Appendix~\ref{FairMVC_efficiency}, we observe that although the contrastive regularizer outperforms the non-contrastive regularizer, the time complexity of the contrastive regularizer is quadratic with respect to the number of instances, whereas the time complexity of non-contrastive regularizer is linear.

\section{Conclusion}
In this paper, we propose \method, a deep fairness-aware multi-view clustering method. \method\ 
maximizes the mutual agreement of the soft membership assignment based on each view; in the meanwhile, it enforces the fairness constraint by requiring that the fraction of different groups in each cluster be approximately the same as the fraction in the whole data set. In addition, we adopt the idea of contrastive learning and non-contrastive learning, and propose novel regularizers to handle heterogeneous data in complex scenarios with missing or noisy features. The experimental results on both synthetic and real-world data sets demonstrate the effectiveness of the proposed framework. We also provide insights regarding the relative performance of contrastive and non-contrastive regularizers in different scenarios.
\section*{Acknowledgment}
This work is supported by National Science Foundation under Award No. IIS-1947203, IIS-2117902, IIS-2137468, IIS-19-56151, IIS-17-41317, and IIS 17-04532, the C3.ai Digital Transformation Institute, MIT-IBM Watson AI Lab, and IBM-Illinois Discovery Accelerator Institute - a new model of an academic-industry partnership designed to increase access to technology education and skill development to spur breakthroughs in emerging areas of technology. The views and conclusions are those of the authors and should not be interpreted as representing the official policies of the funding agencies or the government.

\bibliographystyle{abbrv}
\small{\bibliography{reference}}

\newpage
\appendix
\section{Appendix}



\subsection{Experimental Setup}
\label{FairMVC_setup}
We mainly evaluate our proposed algorithm on three data sets with fairness constraints, including Credit card clients data set~\footnote{\url{https://archive.ics.uci.edu/ml/datasets/default+of+credit+card+clients}}, Bank Marketing Data set~\footnote{\url{https://ashryaagr.github.io/Fairness.jl/dev/datasets/\#Bank-Marketing-Dataset}}, Zafar data set~\footnote{\url{https://ashryaagr.github.io/Fairness.jl/dev/datasets/\#Fairness.genZafarData}}, and two data sets without fairness constraints, which are Noisy MNIST data set~\footnote{\url{http://yann.lecun.com/exdb/mnist/}}, and X-ray Microbeam (XRMB) ~\footnote{\url{https://ttic.uchicago.edu/~klivescu/XRMB\_data/full/README}}.
Specifically, the Credit card clients data set describes the customers' default payments in Taiwan and this data set consists of 30,000 samples with 24 attributes. The sensitive feature in this data set is gender. Bank Marketing Data set is associated with direct marketing campaigns of a Portuguese banking institution and it aims to see if the product (bank term deposit) would be (`yes') or not (`no') subscribed. The sensitive feature in this data set is marital status. This data set consists of 1,000 instances and 20 attributes. Zafar data set~\cite{DBLP:conf/aistats/ZafarVGG17} is a widely-used synthetic data set, where one binary value is generated as the sensitive feature. For the Credit Card, and Bank Marketing data set, we use two non-linear functions (\eg, $Sigmoid$ and $Relu$) to generate two views.  Noisy MNIST data set originally consists of 70,000 images of handwritten digits and we follow~\cite{WangALB15} by adding white Gaussian noise to each pixel to generate the first view, and randomly rotating a figure with an angle from [-$\frac{\pi}{4}$, $\frac{\pi}{4}$] to generate the second view. XRMB~\cite{westbury1994x} is a multi-view multi-class data set, which consists of 40 binary labels and two views. The first view is acoustic data with 273 features and the second view is articulatory data with 112 features.  As some state-of-the-art methods are very slow, we reduce the number of instances for some large data sets to ensure that we can include the results of most baselines. We randomly sample 5,000 instances from the Noisy MNIST data set and 3,000 instances from the XRMB data set from 6 classes.

\noindent\textbf{Configuration:}
In all experiments, we set the learning rate to 0.001 and the weight decay rate to be 0.0005.
The optimizer is momentum SGD.  The neural network structure for $f^v(\cdot)$ of the proposed methods is a two-layer Multi-layer Perceptron (MLP) and The neural network structure for $g^v(\cdot)$ of the proposed methods is a one-layer MLP. The experiments are repeated 5 times if not specified. The code of our algorithms could be found in the link~\footnote{\url{https://github.com/Leo02016/FairMVC}}. The experiments are performed on a Windows machine with a 16GB RTX 5000 GPU and 64GB memory.

\subsection{Experimental results on real-world data sets without sensitive features}
\label{appendix_no_fairness}
In this subsection, we evaluate the performance of our proposed method on two datasets without sensitive features, including Noisy MNIST and XRMB. Specifically, we randomly sample 5,000 instances from the Noisy MNIST data set and 3,000 instances from the XRMB data set from 6 classes.
Table~\ref{table:non_fairness} shows the performance of state-of-the-art methods and our proposed methods. By observations, we find that our proposed method \method-N and \method-C outperform all state-of-the-art methods on the XRMB data set and Noisy MNIST data set in Table~\ref{table:non_fairness}.

\begin{table}
\centering
\caption{Results on real-world data sets without sensitive features}
\resizebox{\columnwidth}{!}{%
\begin{tabular}{|*{3}{c|}}
\hline   -    & \textbf{NMI}     & \textbf{NMI}   \\
\hline \textbf{Model}  & XRMB & Noisy MNIST \\
\hline K-means                  & 0.1692 $\pm$ 0.0049 &  0.3882 $\pm$ 0.0117   \\   
\hline DEC                      & 0.2012 $\pm$ 0.0086 &  0.4899 $\pm$ 0.0227   \\    
\hline CC                       & 0.2107 $\pm$ 0.0100 &  0.4902 $\pm$ 0.0101   \\   
\hline MvDSCN                   & 0.2056 $\pm$ 0.0078 &  0.4770 $\pm$ 0.0128   \\   
\hline \method-C                & 0.2163 $\pm$ 0.0101 &  \textbf{0.4997  $\pm$ 0.164 }  \\   
\hline \method-N                & \textbf{0.2214 $\pm$ 0.0071} & 0.4686 $\pm$ 0.182 \\   
\hline
\end{tabular}
}
\label{table:non_fairness}
\end{table}

\subsection{Efficiency Analysis}
\label{FairMVC_efficiency}
\begin{figure}
\caption{Efficiency analysis: number of instances vs running time (Best viewed in color)}
\begin{center}
\includegraphics[width=0.65\linewidth]{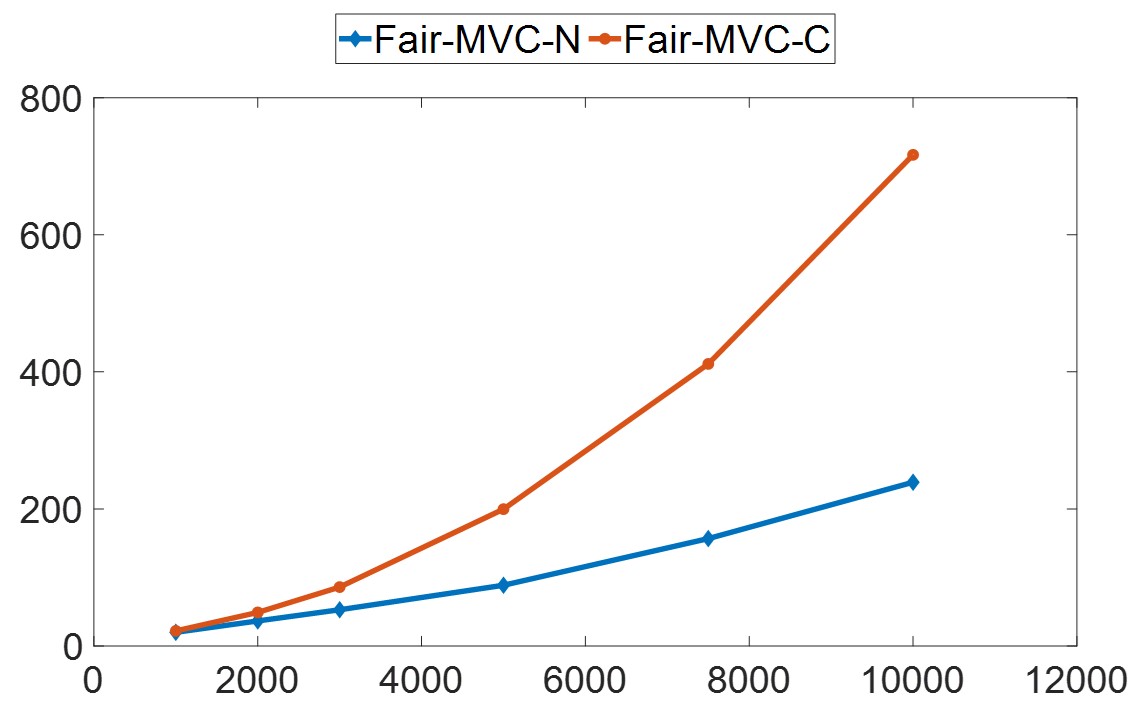} \\
\end{center}
\label{fig_efficiency_analysis}
\end{figure}

In this subsection, we analyze the efficiency of our proposed algorithm with two different regularization terms (\ie, contrastive regularizer and non-contrastive regularizer) on the Zafar data set. Specifically, we increase the number of samples from 1,000 to 10,000 and record the running time (in seconds) for these two regularizations in Figure~\ref{fig_efficiency_analysis}. The total number of iterations is 1000. The x-axis of this figure is the number of samples and the y-axis is the running time. By observations, we find that the running time is almost linear to the number of samples for non-contrastive regularizer (\ie, \method-N) and the running time is quadratic to the number of samples for contrastive regularizer (\ie, \method-C). The reason is that for \method-C, we need to compute the similarity of any given two samples in the denominator of contrastive regularizer in Equation \ref{l_ctr}, while \method-N only computes the similarity of two views for the same sample. Thus, the time complexity of contrastive-based regularization is $O(n^2)$, whereas the time complexity of non-contrastive-based regularization is $O(n)$, where $n$ is the number of samples.

\begin{figure}[h]
\caption{Parameter analysis on Credit Card data set for \method-C }
\begin{center}
\begin{tabular}{cc}
\hspace{-0.5cm}
\includegraphics[width=0.45\linewidth]{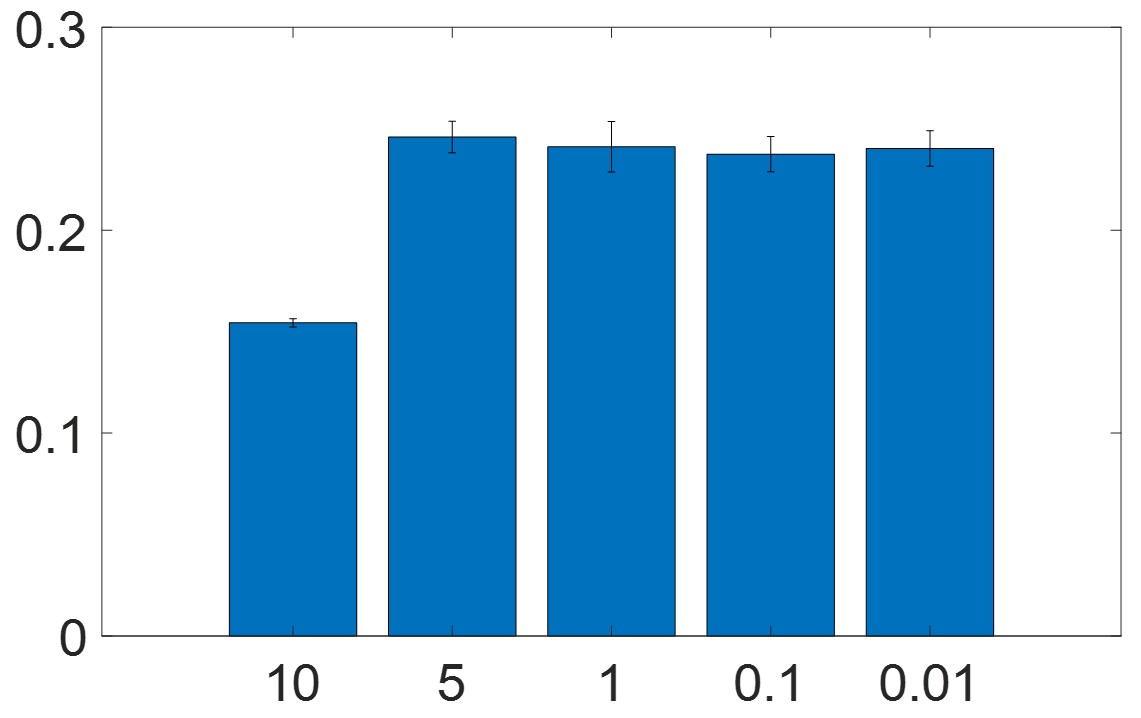} &
\hspace{-0.5cm}
\includegraphics[width=0.45\linewidth]{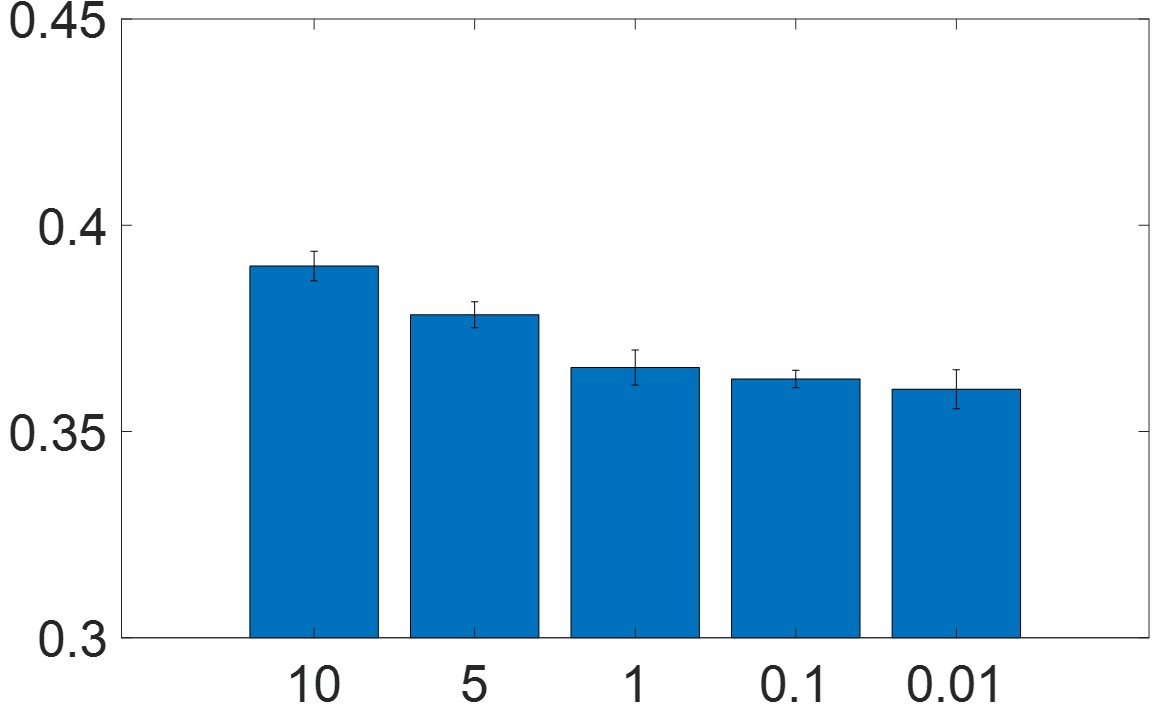} \\
(a) $\alpha$ vs NMI &
(b) $\alpha$ vs balance score \\
\hspace{-0.5cm}
\includegraphics[width=0.45\linewidth]{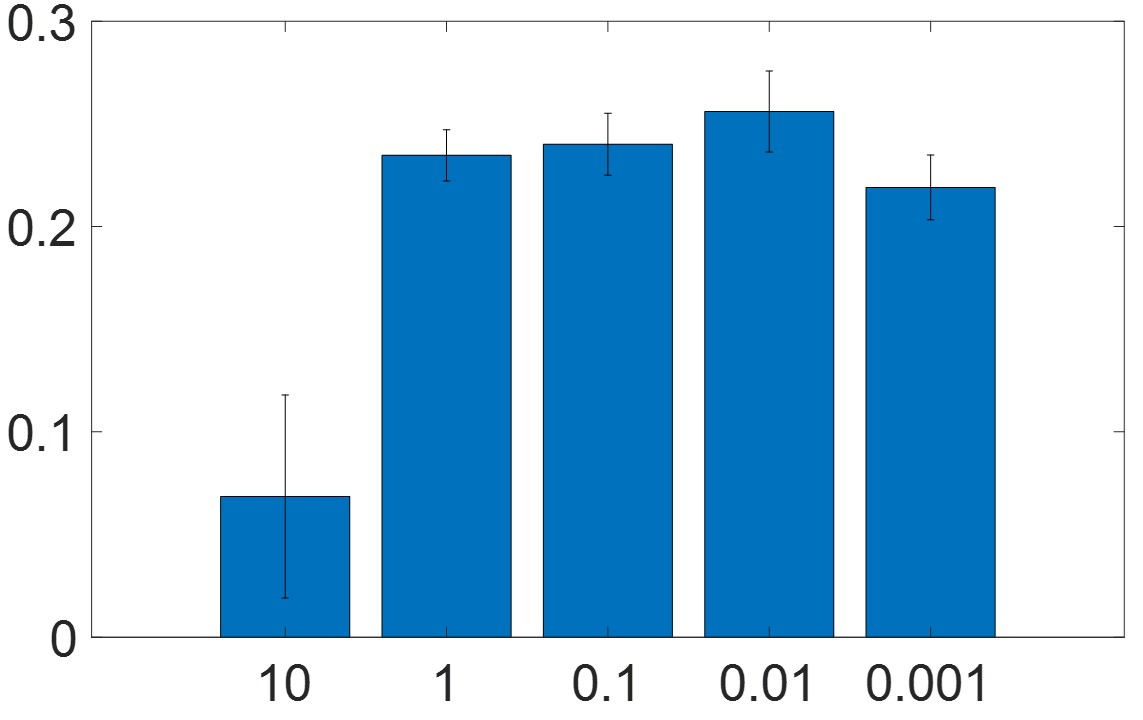} &
\hspace{-0.5cm}
\includegraphics[width=0.45\linewidth]{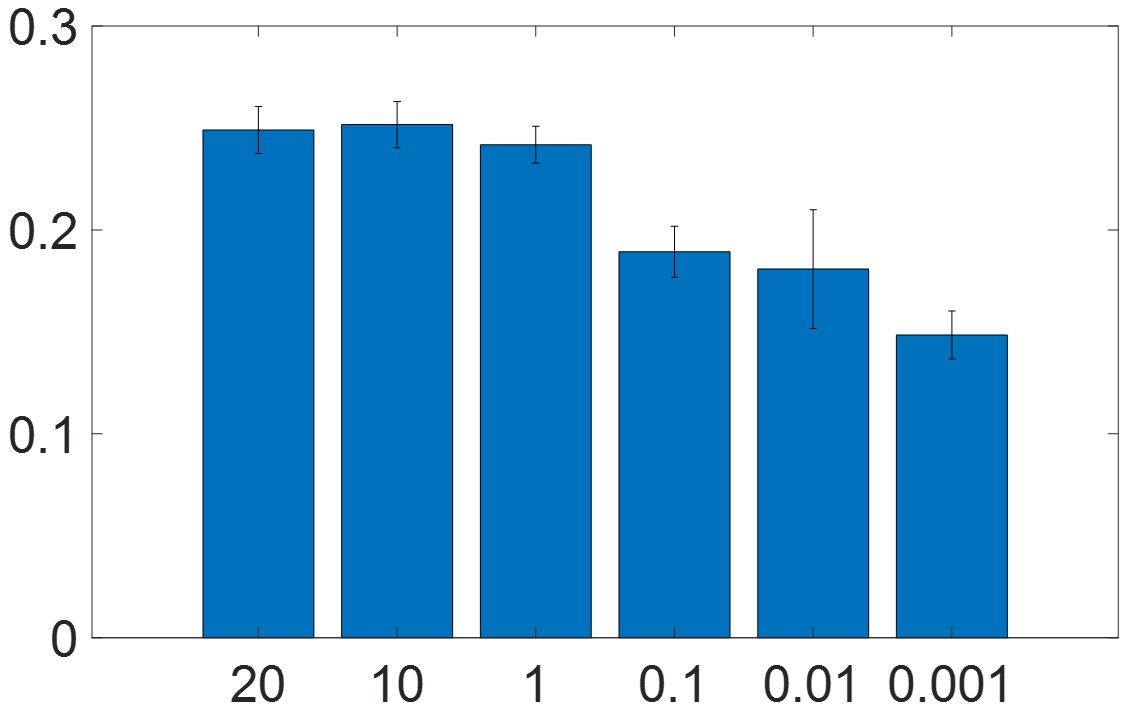} \\
(c) $\beta$ vs NMI &
(d) $\gamma$ vs NMI \\
\end{tabular}
\end{center}
\label{fig_parameter_analysis}
\end{figure}

\begin{figure}[h]
\caption{Parameter analysis on Credit Card data set for \method-N}
\begin{center}
\begin{tabular}{cc}
\hspace{-0.5cm}
\includegraphics[width=0.49\linewidth]{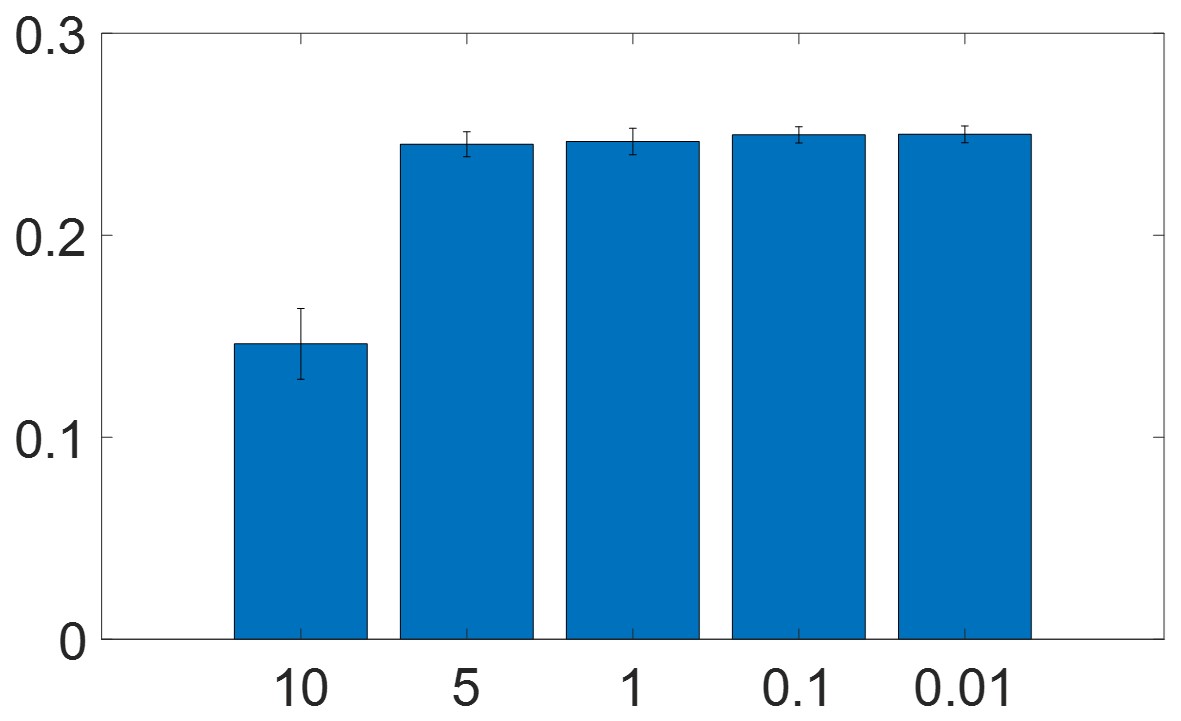} &
\hspace{-0.5cm}
\includegraphics[width=0.49\linewidth]{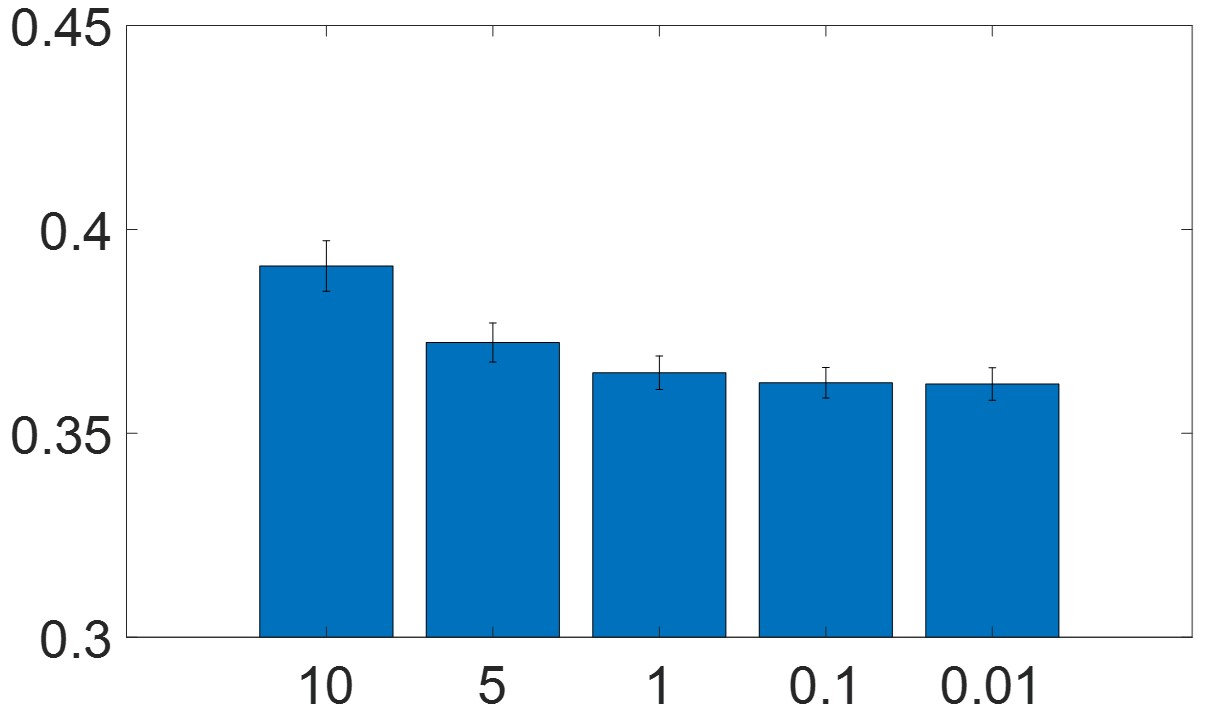} \\
(a) $\alpha$ vs NMI &
(b) $\alpha$ vs balance score \\
\hspace{-0.5cm}
\includegraphics[width=0.49\linewidth]{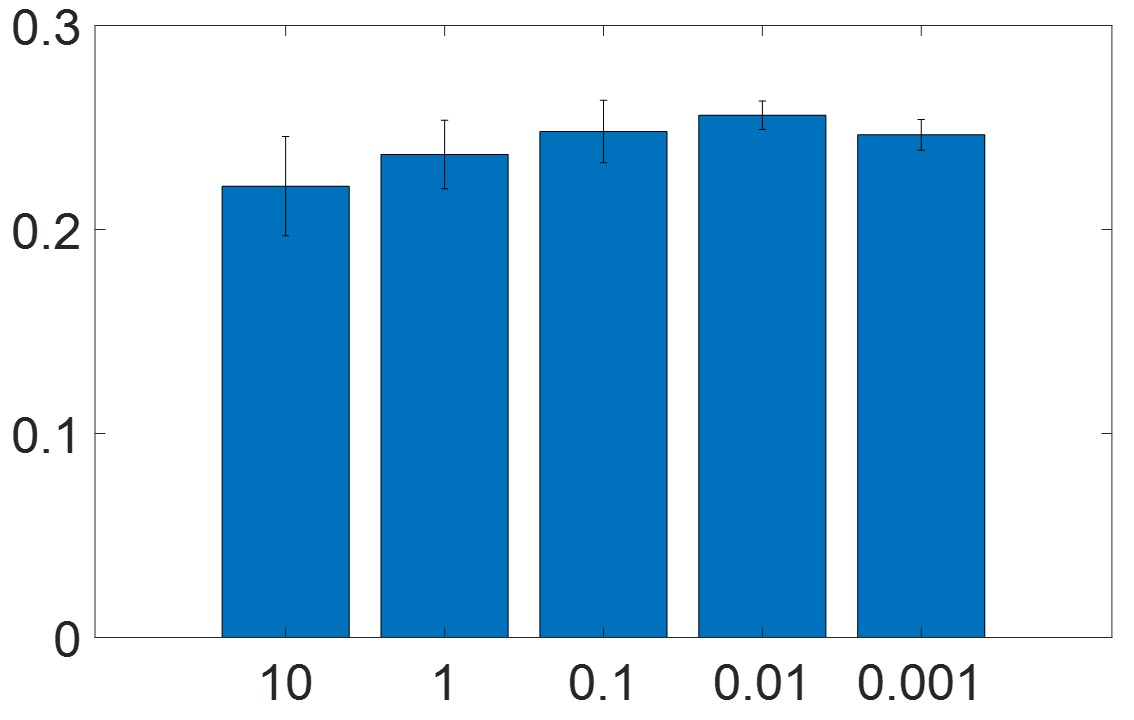} &
\hspace{-0.5cm}
\includegraphics[width=0.49\linewidth]{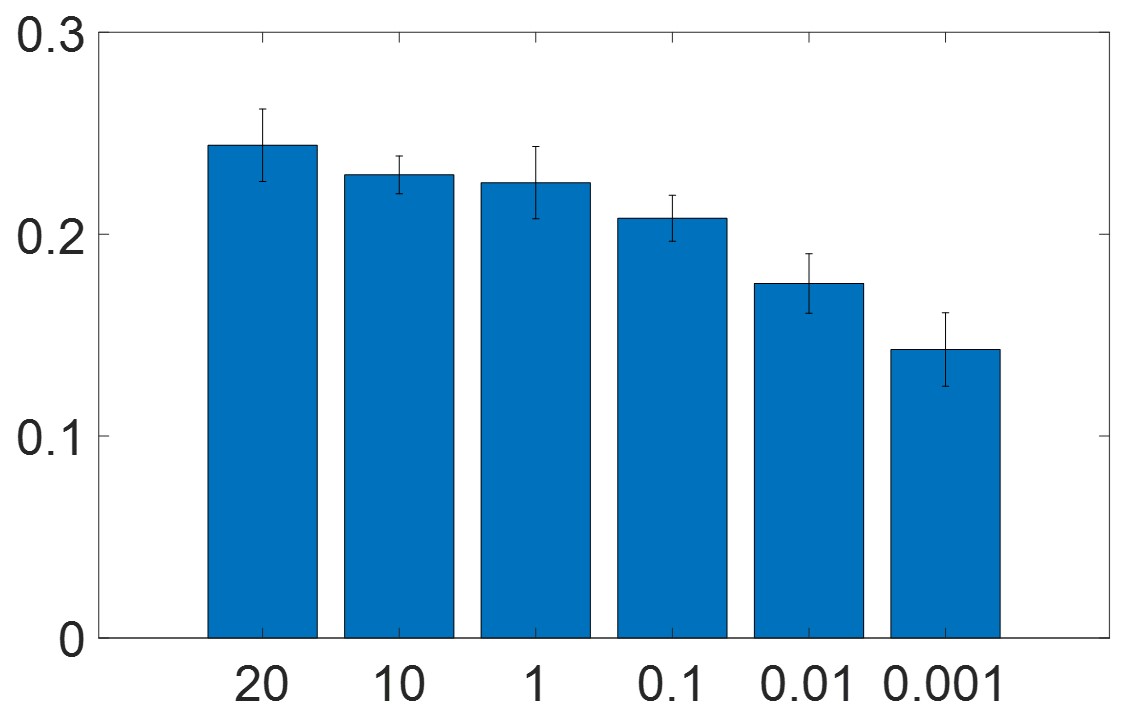} \\
(c) $\beta$ vs NMI &
(d) $\gamma$ vs NMI \\
\end{tabular}
\end{center}
\label{fig_parameter_analysis_2}
\end{figure}

\subsection{Parameter Analysis}
\label{FairMVC_parameter_analysis}
In the subsection, we conduct the parameter analysis regarding $\alpha$, $\beta$, and $\gamma$ for \method-C. Specifically, we change the value of one hyper-parameter, fix the other hyper-parameters, and report the results. Figure~\ref{fig_parameter_analysis} shows the results regarding these three hyper-parameters. Figure~\ref{fig_parameter_analysis} (a) and Figure~\ref{fig_parameter_analysis} (b) show the NMI and balance score when we change the value of $\alpha$. We observe that when $\alpha=10$, \method-C achieves the highest balance score but its performance is the worst as the algorithm mainly focuses on minimizing the fairness loss. When we reduce the value of $\alpha$, then the performance increases to 24.5\% at $\alpha=5$ and starts to change slightly from $\alpha=5$ to $\alpha=0.01$. However, Figure~\ref{fig_parameter_analysis} (b) shows that the results gradually become unfair (with a lower balance score) if we decrease the value of $\alpha$ from 10 to 0.01. Based on these observations, we may conclude that there is a trade-off between the balance score and NMI, and \method-C tends to have a higher NMI and a lower balance score with a lower $\alpha$ and vice versa. Figure~\ref{fig_parameter_analysis} (c) shows the performance of \method-C by changing the value of $\beta$. We observe that the algorithm achieves the best performance when $\beta=0.01$ and it tends to have a large standard deviation when $\beta$ is large (\eg, $\beta=10$). In the overall objective function (\eg, Equation ~\ref{overall}), $\beta$ is the weight of the contrastive regularizer. A large number of $\beta$ greatly reduces the importance of other components (\eg, the centrality of the clustering) and thus it leads to the unstable performance of clustering results. Figure~\ref{fig_parameter_analysis} (d) shows the performance of \method-C with different value of $\gamma$. We observe that the algorithm achieves the best performance when $\gamma$ is around 10. In the overall objective function (\eg, Equation ~\ref{overall}), $\gamma$ controls the importance of centrality and a large value of $\gamma$ implies that the instances assigned to the same cluster will be closer in the hidden space. Thus, in Figure~\ref{fig_parameter_analysis} (d), \method-C with a large value of $\gamma$ usually tends to have a better performance.

Next, we conduct the parameter analysis regarding $\alpha$, $\beta$, and $\gamma$ for \method-N. Specifically, we change the value of one hyper-parameter, fix the rest hyper-parameters, and report the performance.  
Figure~\ref{fig_parameter_analysis_2} shows the results regarding these three hyper-parameters. Figure~\ref{fig_parameter_analysis_2} (a) and Figure~\ref{fig_parameter_analysis_2} (b) show the NMI and balance score when we change the value of $\alpha$. We observe that when $\alpha=10$, \method-N achieves the best balance score; when we decrease $\alpha$, then the performance increases but the results become unfair (with a lower balance score). Based on the results from Figure~\ref{fig_parameter_analysis_2} (a) and Figure~\ref{fig_parameter_analysis_2} (b), we can also conclude that there is a trade-off between the balance score and NMI, and \method-N has a higher NMI with a lower balance score with smaller $\alpha$.
Figure~\ref{fig_parameter_analysis_2} (c) shows the performance of \method-N by changing the value of $\beta$. We observe that the algorithm achieves the best performance when $\beta=0.01$ and it tends to have a large standard deviation when $\beta$ is large. In the overall objective function (\eg, Equation ~\ref{overall}), $\beta$ is the weight of the contrastive regularizer. A large number of $\beta$ greatly reduce the importance of other components (\eg, the centrality of the clustering), and thus it leads to the unstable performance of clustering results.
Figure~\ref{fig_parameter_analysis_2} (d) shows the performance of \method-N with different value of $\gamma$. We observe that the algorithm achieves the best performance when $\gamma$ is 20. In the overall objective function (\eg, Equation ~\ref{overall}), $\gamma$ controls the importance of centrality and a large value of $\gamma$ implies that the instances assigned to the same cluster will be closer in the hidden space. Thus, in Figure~\ref{fig_parameter_analysis_2} (d), \method-N with a large value of $\gamma$ usually tends to have a better performance.

\balance

\end{document}